\documentclass[conference]{IEEEtran}
\IEEEoverridecommandlockouts
\usepackage{cite}
\usepackage{amsmath,amssymb,amsfonts}
\usepackage{algorithmic}
\usepackage{graphicx}
\usepackage{textcomp}
\usepackage{xcolor}
\usepackage{bm}
\usepackage{hyperref}
\usepackage{booktabs}
\def\BibTeX{{\rm B\kern-.05em{\sc i\kern-.025em b}\kern-.08em
    T\kern-.1667em\lower.7ex\hbox{E}\kern-.125emX}}
\begin{document}

\title{Real-Time ESFP: Estimating, Smoothing, Filtering, and     Pose‑Mapping
}

\author{
\IEEEauthorblockN{Qifei Cui\IEEEauthorrefmark{1}, Yuang Zhou\IEEEauthorrefmark{1}, Ruichen Deng\IEEEauthorrefmark{1}\thanks{\IEEEauthorrefmark{1} All authors contributed equally.}}
\IEEEauthorblockA{\textit{School of Engineering and Applied Science}\\
\textit{University of Pennsylvania}\\
Philadelphia, USA \\
\{qifei@seas.upenn.edu, yuangzho@seas.upenn.edu, ruichend@seas.upenn.edu\}\\
\href{Github: }{https://github.com/Qifei-C/Genuine-ESFP}
}
}

\maketitle

\begin{abstract}
This paper presents ESFP, an end-to-end pipeline that converts monocular RGB video into executable joint trajectories for a low-cost 4-DoF desktop arm. ESFP comprises four sequential modules. (1) Estimating: ROMP lifts each frame to a 24-joint 3-D skeleton. (2) Smoothing: the proposed HPSTM—a sequence-to-sequence Transformer with self-attention—combines long-range temporal context with a differentiable forward-kinematics decoder, enforcing constant bone lengths and anatomical plausibility while jointly predicting joint means and full covariances. (3) Filtering: root-normalised trajectories are variance-weighted according to HPSTM’s uncertainty estimates, suppressing residual noise. (4) Pose-Mapping: a geometric retargeting layer transforms shoulder–elbow–wrist triples into the uArm’s polar workspace, preserving wrist orientation. \footnote{See demo vedio at https://www.youtube.com/watch?v=7yrYrcs5UFk}
\end{abstract}

\begin{IEEEkeywords}
3-D human pose estimation, Transformer, Manifold-constrained, Forward kinematics, Vision-to-robot imitation
\end{IEEEkeywords}

\section{Introduction}
\begin{figure}[htbp]
  \centering
  \includegraphics[width=1\linewidth]{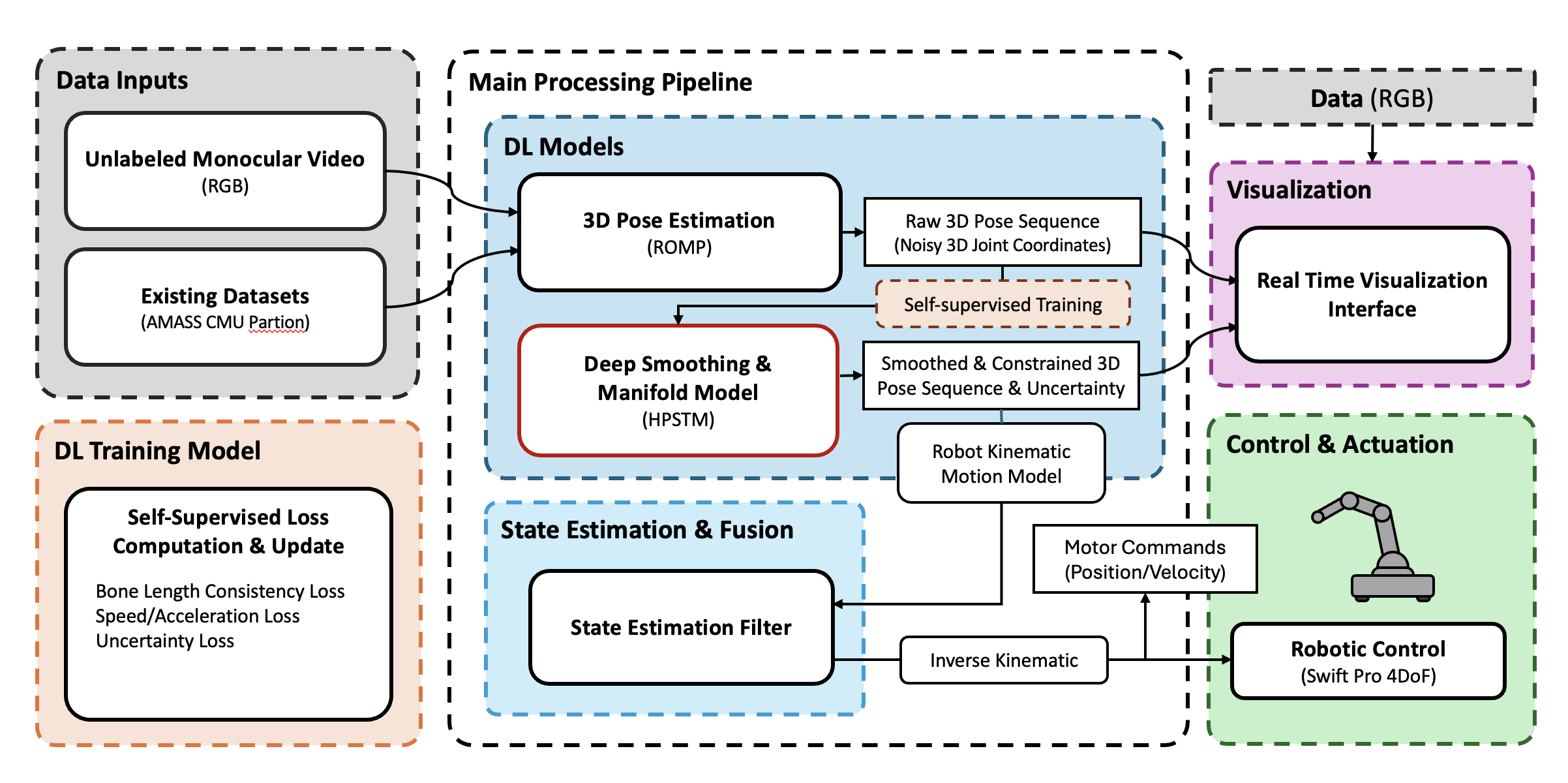}
  \caption{ESFP Workflow}
  \label{fig:frames}
\end{figure}

The estimation of three-dimensional (3D) human pose from various sensors, such as monocular RGB cameras, multi-view camera systems, or depth sensors, is a fundamental task in computer vision. \cite{zhang2023survey} However, the raw 3D pose data obtained from these systems, typically represented as a set of $\mathbb{R}^3$ joint coordinates, are often fraught with imperfections. Monocular 3D human pose estimation, while highly accessible due to its minimal hardware requirements (requiring only a single camera), is particularly susceptible to inherent challenges like depth ambiguity, where multiple distinct 3D poses can project to the same 2D image, and self-occlusion, where parts of the body obscure others. \cite{zhang2023survey} These difficulties are compounded by the limited diversity of large-scale 3D pose datasets, which are often collected in controlled laboratory environments and may not generalize well to "in-the-wild" scenarios with varied backgrounds, lighting conditions, and human appearances. \cite{aaai2020} The prevalence of monocular estimation, driven by its practicality, thus directly contributes to the commonality of these errors, making robust post-processing or integrated refinement essential.
\color{black}

\section{Related Work}
\subsection{SMPL: A Skinned Multi-Person Linear Model}
The SMPL model (\emph{Skinned Multi-Person Linear model}) is a widely adopted statistical representation of the human body that combines a low-dimensional parameter space with linear blend skinning (LBS) to generate realistic, fully differentiable 3-D meshes.  It underpins many state-of-the-art pipelines for monocular pose estimation, motion capture, and animation because it offers three essential properties: a compact pose–shape space learned from thousands of laser scans, an articulated skeletal structure compatible with traditional skinning, and analytic gradients with respect to both pose and shape parameters\,\cite{SMPL2015}. SMPL models the human mesh can be expressed as $(\bm\beta, \bm\theta, \mathbf{T}, \mathbf{J})$ where $\bm\beta\in\mathbb{R}^{10}$ encodes personalized shape (a latent representation),
$\bm\theta\in\mathbb{R}^{24\times3}$ stores the joint rotations,
$\mathbf{T}(\bm\beta,\bm\theta)$ is the posed template,
$\mathbf{J}(\bm\beta)$ are the pose-dependent joint locations.

\subsection{ROMP: Regress Once, Multiple Person}
\begin{figure}
    \centering
    \includegraphics[width=0.6\linewidth]{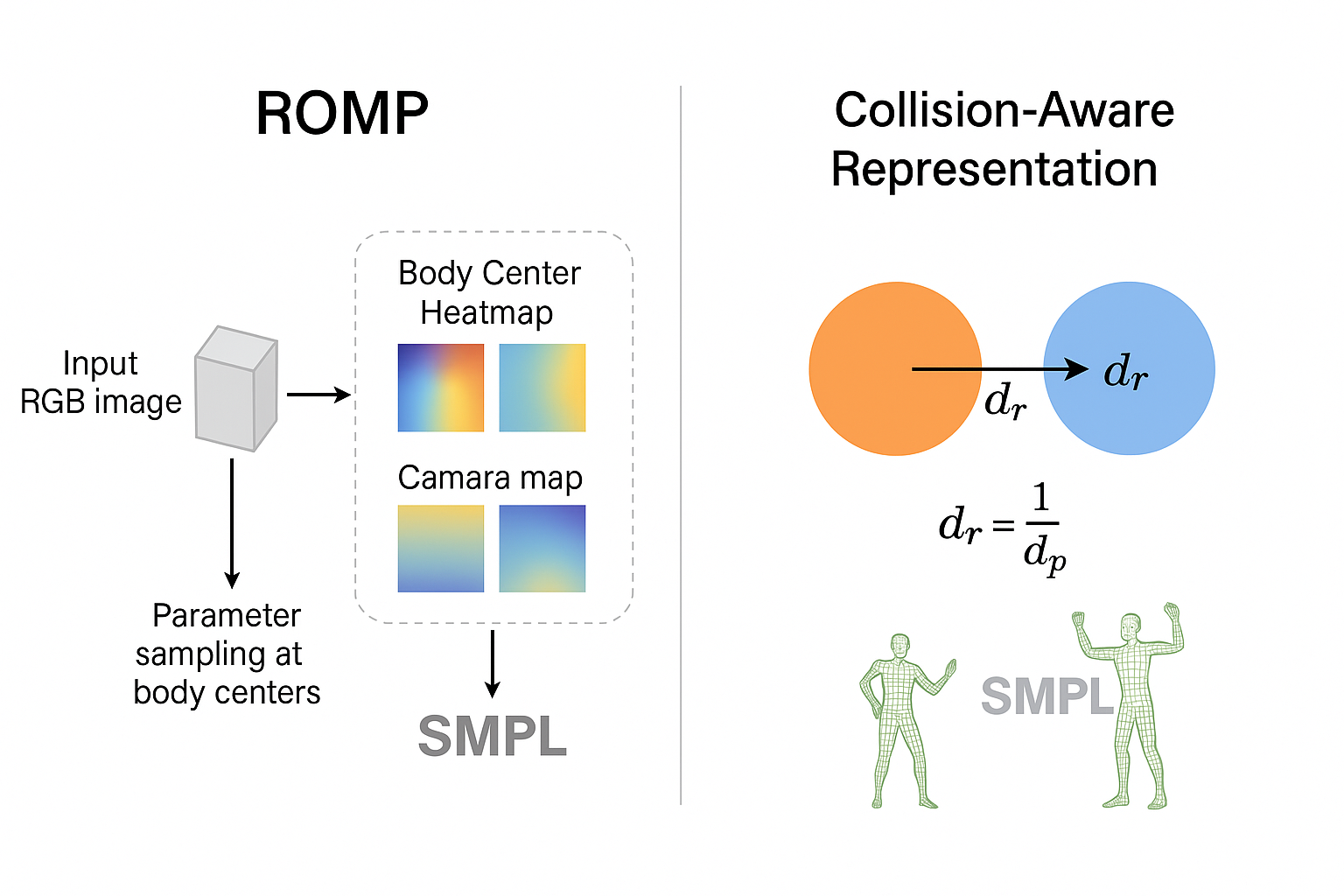}
    \caption{ROMP pipeline.}
    \label{fig:romp}
\end{figure}

Monocular 3-D human pose estimation has witnessed rapid progress since the introduction of parametric body models, e.g. SMPL. Early pipelines of multi-person 3-D pose estimation involves equipping a single-person estimator with 2-D person detector such as YOLO~\cite{Jiang2020, VIBE2020}. More modern works has moved towards end-to-end architectures that regress pose and camera parameters directly from pixels, either for single persons~\cite{HMR2018,SPIN2019} or within cropped regions produced by object detectors for multiple persons~\cite{MHFormer2022}.  Although cropping simplifies scale variation, it fragments global context and introduces expensive per-instance forward passes.
Multi-person settings exacerbate depth ordering and occlusion issues.  Collision-aware or center-based representations have been proposed for object detection~\cite{CenterNet2019}, inspiring holistic formulations that avoid region proposals.  ROMP (\emph{Regress Once, Multiple People})~\cite{ROMP2021} pushes this philosophy further by predicting, at \emph{every} image location, both a body-center confidence and the full SMPL$+$camera parameter vector, thereby removing explicit instance separation while remaining single-shot and fully convolutional.
ROMP turns multi-person mesh recovery into dense prediction on a single feature map produced by a backbone with two coordinate channels.  Three lightweight heads transform the \(128\times128\) features into a body-center heat-map \(C_m\), a weak-perspective camera map \(A_m\) and an SMPL parameter map \(S_m\); stacking the last two yields, for every pixel, a \(145\)-D vector containing scale \(s\), translations \((t_x,t_y)\), ten shape coefficients \(\bm\beta\) and 71 3-D joint locations, where 71 joints comprises a refined version of the SMPL body model.  Ground-truth centers are encoded as Gaussians whose kernel adapts to person size
\begin{equation}
  k = k_l + \Bigl(\tfrac{d_{bb}}{\sqrt{2}\,W}\Bigr)^{2} k_r ,
\end{equation}
while a \emph{collision-aware representation} (CAR) repels overlapping peaks. Figure~\ref{fig:romp} illustrates the full pipeline.

\subsection{Pose Smoothing and Manifold Learning}\label{ssec:related:smooth}
Video-level refinement seeks to suppress the high-frequency jitter that afflicts frame-wise estimators.  
\textbf{SmoothNet}~\cite{Zeng2022SmoothNet} learns per-joint fully connected networks over short temporal windows; although effective, it ignores inter-joint correlations.  
\textbf{FLK}~\cite{Martini2024FLK} combines a Kalman filter with a learned GRU motion prior, but still treats joints independently and does not enforce strict limb-length constraints.  
Manifold-based approaches, e.g.\ ManiPose~\cite{Rommel2024ManiPose}, constrain hypotheses to lie on a fixed-length kinematic manifold, yet they address multi-hypothesis estimation rather than dedicated smoothing.  
Our HPSTM module (Sec.~\ref{sec:hpstm}) unifies long-range temporal attention with an explicit forward-kinematics manifold, yielding smoother and anatomically valid trajectories.

\section{Approach}
\subsection{ROMP: A First Glance of Human Pose}
For every RGB frame $\mathbf{I}_t\in\mathbb{R}^{H\times W\times 3}$ we obtain a raw, frame-wise estimate of the body configuration by passing the image through a frozen ROMP network, 
\begin{equation}
  \bigl(\hat{s}_t,\hat{t}_{x,t},\hat{t}_{y,t},\,
        \hat{\mathbf{J}}^{\mathrm{c}}_t\bigr)
  \;=\;
  F_{\boldsymbol\phi}(\mathbf{I}_t),
  \label{eq:romp_core}
\end{equation}
where $F_{\boldsymbol\phi}$ denotes ROMP’s convolutional backbone and prediction heads (weights $\boldsymbol\phi$ are fixed),  
\(\hat{s}_t\) and $(\hat{t}_{x,t},\hat{t}_{y,t})$ form a weak-perspective camera, and  
\(\hat{\mathbf{J}}^{\mathrm{c}}_t\in\mathbb{R}^{24\times 3}\) contains the first twenty-four SMPL joint coordinates in the model’s canonical (root-centered) space. No mesh vertices, shape coefficients or auxiliary joints are retained.
\vspace{0.3em}
\noindent\textbf{Camera-centric lifting.}
To express joints in the actual camera frame we undo the weak-perspective projection by
\begin{equation}
  \mathbf{J}_t
  \;=\;
  \frac{1}{\hat{s}_t}\,
  \mathbf{K}^{-1}
  \Bigl(
    \hat{\mathbf{J}}^{\mathrm{c\top}}_{t,xy}
    \;+\;
    \mathbf{1}\otimes
    [\hat{t}_{x,t},\hat{t}_{y,t}]
  \Bigr)^{\!\top},
  \label{eq:unproject}
\end{equation}
where  
\(\hat{\mathbf{J}}^{\mathrm{c}}_{t,xy}\) denotes the $x$–$y$ sub-matrix of the canonical joints,  
\(\mathbf{K}\) is the intrinsic calibration matrix of our capture camera, and the division by \(\hat{s}_t\) provides an approximate depth.  
The result is a tensor \(\mathbf{J}_t\in\mathbb{R}^{24\times 3}\) that serves as the sole input to the subsequent temporal smoothing stage.

\subsection{HPSTM: Transformer-Based Smoothing Module}\label{sec:hpstm}

The proposed \textbf{HPSTM} refines noisy ROMP outputs into smooth, physically plausible sequences suitable for robot control. \footnote{\url{https://github.com/Qifei-C/HPSTM}}  
It employs an encoder–decoder Transformer whose multi-head self-attention captures \emph{long-range temporal dependencies} and \emph{cross-joint correlations}, unlike window-bound MLP smoothers such as SmoothNet~\cite{Zeng2022SmoothNet} or per-joint Kalman variants like FLK~\cite{Martini2024FLK}.

\medskip
\noindent\textbf{Architecture.}  
Given a window of $T$ frames ($T\!=\!31$ in our implementation), the encoder aggregates spatio-temporal features, while the decoder—driven by learned query embeddings—predicts (i) root translation, (ii) joint rotations in quaternion form, and (iii) per-joint bone lengths.  
This sequence-to-sequence design allows HPSTM to non-causally smooth each frame using context from both past and future, surpassing the causal GRU prior in FLK.

\medskip
\noindent\textbf{Forward-kinematics manifold.}  
Rather than outputting Cartesian joint positions directly, HPSTM passes predicted pose parameters through a differentiable forward-kinematics (\textsc{fk}) layer that reconstructs global joint coordinates.  
Because limb lengths are constrained by the learned positive bone-length vector, every output pose lies \emph{on the human kinematic manifold}, eliminating the limb-stretch artifacts occasionally observed with SmoothNet and FLK.  
This strategy generalizes the manifold constraints used for hypothesis selection in ManiPose~\cite{Rommel2024ManiPose} to a fully differentiable smoothing network.

\medskip
\noindent\textbf{Probabilistic output.}  
HPSTM additionally predicts a full $3\times3$ covariance matrix for each joint at every frame and is trained with a negative log-likelihood loss.  
Neither SmoothNet nor FLK provides such uncertainty quantification; ManiPose addresses ambiguity via multiple hypotheses but lacks an explicit confidence measure.  
Covariance estimates enable downstream controllers to modulate motion speed or compliance based on predicted reliability.

\medskip
\noindent\textbf{Training losses.}  
The model is optimized end-to-end with a weighted sum of (i) $\ell_{1}$ joint-position error, (ii) bone-length consistency, and (iii) Gaussian negative log-likelihood.  
No additional jerk or acceleration regularizer is required—the Transformer’s attention already suppresses high-frequency noise.

In summary, HPSTM combines \emph{global temporal attention}, \emph{explicit kinematic constraints}, and \emph{learned uncertainty} to deliver refined pose sequences that are smoother, anatomically valid, and reliability-aware—features critical for robust vision-to-robot imitation.

\medskip
\noindent\textbf{Training Curriculum}
\label{sec:training}

To ensure stable convergence and calibrated uncertainty estimation, HPSTM is trained following a \emph{three-stage curriculum}:

\textit{Stage~1: Manifold Pre-training.}
The network is initially exposed to clean AMASS sequences. During this stage, the covariance prediction head is disabled. Optimization is performed solely with the position loss $\mathcal{L}_{\text{pos}}$ which is to minimize the discrepancy between the predicted 3D joint positions $\hat{\mathbf{P}}$ and the ground truth positions $\mathbf{P}^*$:
\begin{equation}
    \mathcal{L}_{\text{pos}}(\hat{\mathbf{P}}, \mathbf{P}^*) = \frac{1}{BSJD} \sum_{b,s,j,d} (\hat{p}_{b,s,j,d} - p^*_{b,s,j,d})^2
\label{eq:pos_loss}
\end{equation}

An AdamW optimizer is employed with an initial learning rate (e.g., $1 \times 10^{-4}$), managed by a \texttt{ReduceLROnPlateau} scheduler based on validation loss. This pre-training encourages the encoder-decoder stack to learn a robust and smooth representation of the human kinematic manifold before encountering corrupted input data.

\textit{Stage~2: Noise-aware Refinement.}
Initializing from Stage~1 weights, the model is then trained on sequences corrupted by a combination of four stochastic perturbations: \mbox{(i) $\mathrm{iid}$ Gaussian} displacement ($\sigma=0.01$\,m),
(ii) bone-length jitter (relative $\sigma=3\%$),
(iii) temporally filtered jitter (signal $\sigma=0.015$\,m, filter window $w=7$),
and (iv) frame-wise joint outliers (probability $0.5\%$, max scale deviation $25\%$ of typical joint range).
The training objective in this stage primarily consists of the position loss $\mathcal{L}_{\text{pos}}$ (with a weight $w_{\text{pos}}=1.0$), augmented with additional penalties: a bone-length consistency loss $\mathcal{L}_{\text{bone}}$ (weighted by $w_{\text{bone}}=0.3$), which implies the deviations of predicted bone lengths $\hat{l}_{b,s,j}$ from their target canonical (rest) lengths\footnote{Here, $l^{*\text{canon}}_{b,j}$ is expanded to match the sequence dimension $S$ for comparison. For the root joint, $l^{*\text{canon}}_{b, \text{root}}$ is typically zero.} $l^{*\text{canon}}_{b,j}$
\begin{equation}
    \mathcal{L}_{\text{bone}}(\hat{\mathbf{L}}_{\text{pred}}, \mathbf{L}^{*\text{canon}}) = \frac{1}{BSJ} \sum_{b,s,j} (\hat{l}_{b,s,j} - l^{*\text{canon}}_{b,j})^2;
\label{eq:bone_loss}
\end{equation} a first-order temporal velocity loss $\mathcal{L}_{\text{vel}}$ (weighted by $w_{\text{vel}}=0.5$), where the velocity for joint $j$ at frame $s$ is defined as $\mathbf{v}_{s,j} = \mathbf{p}_{s+1,j} - \mathbf{p}_{s,j}$; the velocity loss, using the $L_1$ norm, is
\begin{equation}
    \mathcal{L}_{\text{vel}}(\hat{\mathbf{P}}, \mathbf{P}^*) = \frac{1}{BS'JD} \sum_{b,s',j,d} |\hat{v}_{b,s',j,d} - v^*_{b,s',j,d}|;
\label{eq:vel_loss}
\end{equation}

and a second-order temporal acceleration loss $\mathcal{L}_{\text{accel}}$ (weighted by $w_{\text{accel}}=0.5$),
where the acceleration for joint $j$ at frame $s$ is $\mathbf{a}_{s,j} = \mathbf{v}_{s+1,j} - \mathbf{v}_{s,j}$. The acceleration loss, using the $L_1$ norm, is
\begin{equation}
    \mathcal{L}_{\text{accel}}(\hat{\mathbf{P}}, \mathbf{P}^*) = \frac{1}{BS''JD} \sum_{b,s'',j,d} |\hat{a}_{b,s'',j,d} - a^*_{b,s'',j,d}|
\label{eq:accel_loss}
\end{equation}

These weighted additions guide the model to learn anatomically valid and temporally smooth reconstructions from noisy observations. The optimizer and learning rate strategy typically continue from Stage~1, or the optimizer may be re-initialized if specific hyperparameter adjustments are needed for this phase.

\textit{Stage~3: Uncertainty Learning (Fine-tuning).}
With the same noise model active as in Stage~2, the covariance prediction head is enabled, and the network is fine-tuned. The primary loss for this stage is the negative Gaussian log-likelihood $\mathcal{L}_{\text{NLL}}$, which is introduced into the total objective. When predicting uncertainty, the model outputs the Cholesky factor $\hat{\mathbf{L}}_{chol_{b,s,j}}$ for the covariance matrix of each joint's 3D position. The Negative Log-Likelihood (NLL) loss for a multivariate Gaussian distribution is used
\begin{align}
    \mathcal{L}_{\text{NLL}} = \frac{1}{BSJ} \sum_{b,s,j} \Bigg[ & \frac{1}{2} \| \hat{\mathbf{L}}_{chol_{b,s,j}}^{-1} (\mathbf{p}^*_{b,s,j} - \hat{\mathbf{p}}_{b,s,j}) \|_2^2 \nonumber \\
    & + \sum_{k=1}^{D} \log((\hat{L}_{chol_{b,s,j}})_{kk}) \nonumber+ \frac{D}{2}\log(2\pi) \Bigg]
\label{eq:nll_loss}
\end{align}
where $(\hat{L}_{chol_{b,s,j}})_{kk}$ are the diagonal elements of the Cholesky factor $\hat{\mathbf{L}}_{chol_{b,s,j}}$, and the first term is the squared Mahalanobis distance.

The contribution of $\mathcal{L}_{\text{NLL}}$ is weighted by a coefficient $\lambda_{\text{NLL}}=10^{-4}$. This relatively small weighting for $\mathcal{L}_{\text{NLL}}$ allows the model to learn calibrated covariance estimates without significantly compromising the mean accuracy achieved in previous stages. For this fine-tuning phase, the optimizer we re-initialized the AdamW optimizer and the learning rate was set to $1\times 10^{-5}$ to facilitate stable learning of the covariance parameters and careful adaptation of the pre-trained weights.

\subsection{Pose Mapping}
\label{subsec:pose_mapping}
The pose--mapping module in the ESFP (\textbf{E}stimating, \textbf{S}moothing, \textbf{F}iltering, and \textbf{P}ose-Mapping) pipeline is responsible for translating filtered 3-D human arm motion into real-time control commands for a \textit{uArm Swift Pro}. \footnote{\url{https://github.com/Yuang-Zhou/skeleton-to-uarm}} This section formalises the kinematic model, details the implementation, and reports experimental performance.

\medskip
\noindent\textbf{Theoretical Basis and Kinematic Strategy}

\textit{Human arm model.} A human arm is typically represented with seven DoF. Let $\mathbf{p}_{s},\mathbf{p}_{e},\mathbf{p}_{w}\in\mathbb{R}^{3}$ denote the 3-D positions of the shoulder, elbow and wrist obtained from AMASS \cite{Mahmood2019AMASS} motion-capture sequences parameterised by the SMPL body model \cite{Loper2015SMPL}. The wrist pose relative to the shoulder is
\begin{equation}
\label{eq:v_human}
    \mathbf v_{\mathrm{h}} = \mathbf{p}_{w}-\mathbf{p}_{s} .
\end{equation}

\textit{Coordinate frames.} The origin of the human body coordinate system is the pelvis position of the SMPL skeleton, and the origin of the robot arm coordinate system is the center point of the base. Human coordinate ($\mathcal H$) is a right-handed frame and uArm coordinate ($\mathcal R$): is a left-handed coordinate:
\begin{itemize}
  \item \textit{Human} ($\mathcal H$): $+X$ forward, $+Y$ left, $+Z$ up.
  \item \textit{uArm} ($\mathcal R$): $+X$ forward (reach), $+Y$ up, $+Z$ right.
\end{itemize}
The constant rotation mapping $\mathbf R_{\mathcal H\to\mathcal R}\in\mathrm{SO}(3)$ is determined during calibration. Applying this rotation gives
\begin{equation}
    \mathbf v_{\mathcal R} = \mathbf R_{\mathcal H\to\mathcal R}  \mathbf v_{\mathrm{h}}.
\end{equation}

\textit{Dynamic scaling.} Owing to workspace disparity, we introduce a scale factor
\begin{equation}
    \lambda = \frac{L_{\mathrm r}}{L_{\mathrm h}},
\end{equation}
where $L_{\mathrm h}=\|\mathbf p_{e}-\mathbf p_{s}\|+\|\mathbf p_{w}-\mathbf p_{e}\|$ is the human arm length and $L_{\mathrm r}$ is a user-defined target reach in the robot frame. The scaled vector is $\lambda\,\mathbf v_{\mathcal R}$.

\textit{Offset and clipping.} A fixed offset $\mathbf o_{\mathcal R}$ positions the conceptual robot shoulder within the SDK frame. The final target is
\begin{equation}
    \mathbf p_{\text{cmd}} = \mathbf o_{\mathcal R} + \lambda\,\mathbf v_{\mathcal R} = \mathbf o_{\mathcal R} + \frac{L_{\mathrm r}}{L_{\mathrm h}}\,\mathbf R_{\mathcal H\to\mathcal R}  \mathbf v_{\mathrm{h}},
\end{equation}
followed by axis-wise clipping to respect mechanical limits.

\begin{figure}[htbp]
  \centering
  \includegraphics[width=0.5\linewidth]{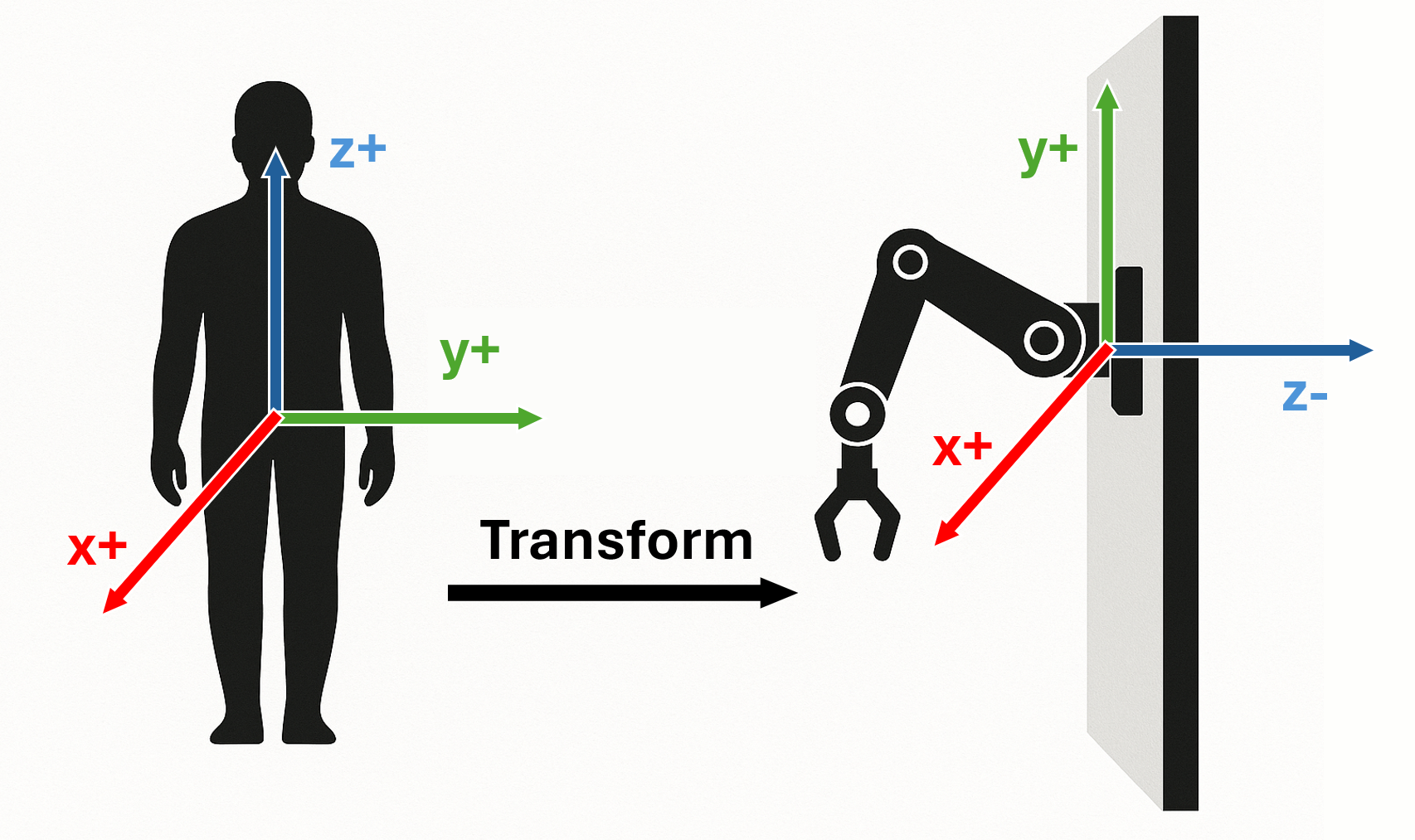}
  \caption{Human ($\mathcal H$) to robot ($\mathcal R$)}
  \label{fig:frames}
\end{figure}

\medskip
\noindent\textbf{Implementation Details}
The algorithm is implemented in our open-source repository.

\begin{enumerate}
  \item \textbf{Data ingestion} — Each AMASS frame is loaded at 30 Hz; noise is reduced with a 1-€ filter.
  \item \textbf{Vector computation} — Eq.~\eqref{eq:v_human} is evaluated per frame.
  \item \textbf{Scaling} — $\lambda$ is updated online; a fallback $\lambda_{0}$ is used when $L_{\mathrm h}<\tau_{\min}$.
  \item \textbf{Mapping} — $\mathbf p_{\text{cmd}}$ is obtained and clipped.
  \item \textbf{Robot actuation} — Non-blocking SDK calls \texttt{set\_position($x,y,z$, speed, wait=False)} achieve 20 Hz command throughput.
\end{enumerate}

\begin{figure}[htbp]
  \centering
  \includegraphics[width=0.6\linewidth]{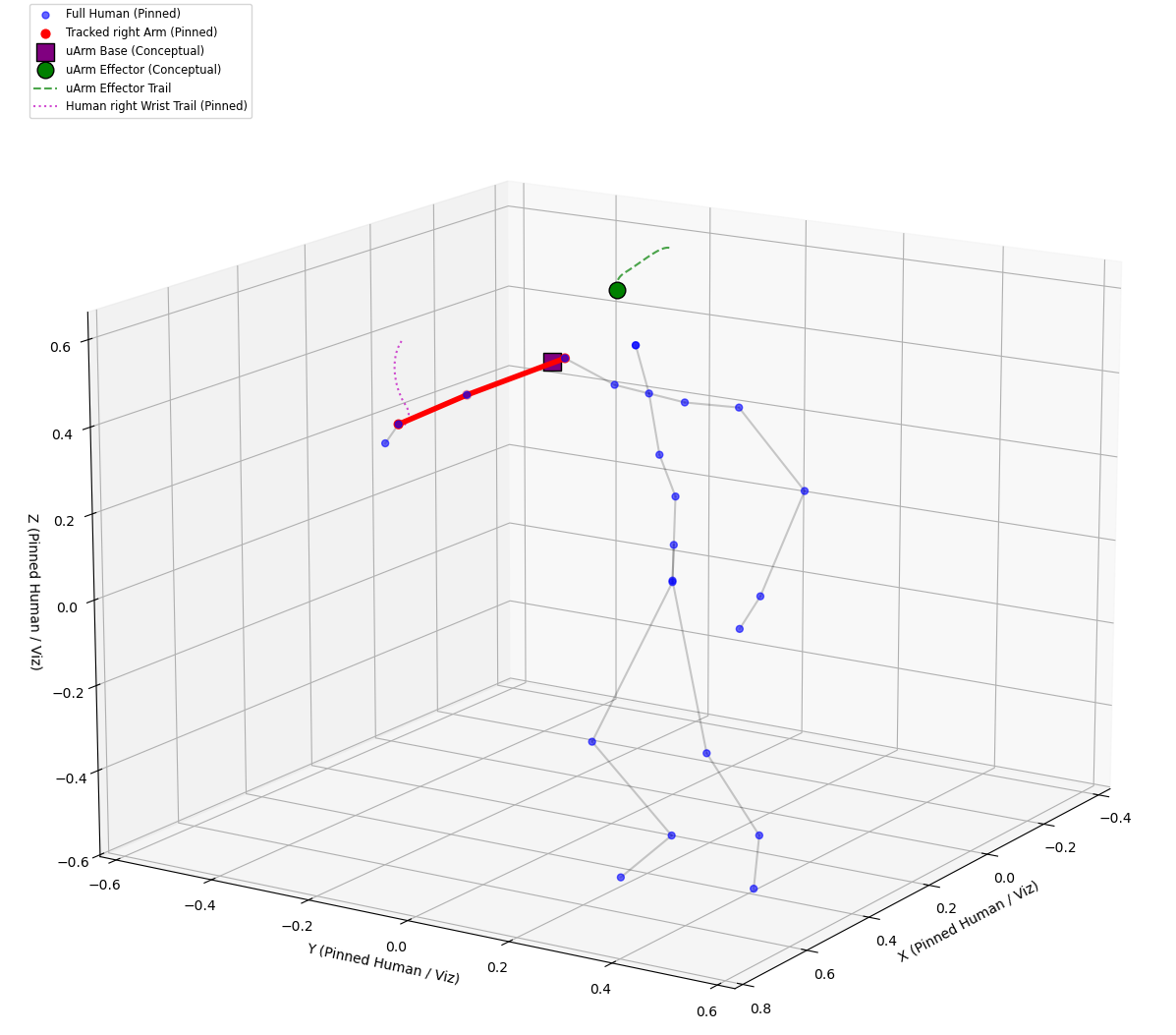}
  \caption{3-D trajectories of human pose (red) vs. robot effector (green) for a reaching motion.}
  \label{fig:traj}
\end{figure}

\begin{figure}
  \centering
  \includegraphics[width=0.5\linewidth]{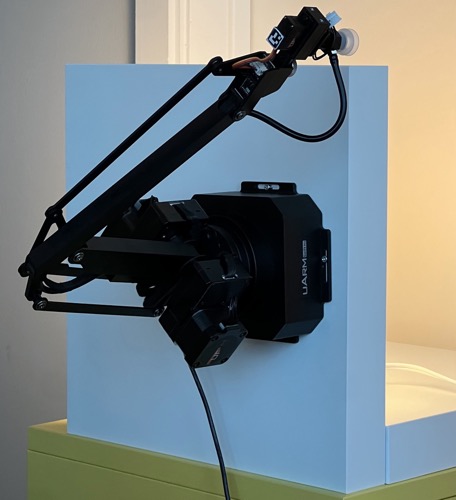}
  \caption{Side-mounted uArm Swift Pro.}
  \label{fig:setup}
\end{figure}

\subsection{Full ESFP Pipeline}

Our proposed Embodied Skeletal Ferrying Persona (ESFP) system achieves real-time human motion mimicry through a modular pipeline. This pipeline integrates three primary stages: (1) initial 3D human pose estimation, (2) probabilistic spatio-temporal pose sequence refinement, and (3) robotic arm mapping and control.

\textbf{Real-time Human Pose Estimation}
The first stage utilizes the \textit{romp tracking} system \cite{ROMP2021} to process an input video stream (live or pre-recorded). For each frame, \textit{romp tracking} performs monocular 3D human pose estimation, outputting a continuous stream of raw 3D coordinates (e.g., a $24 \times 3$ array per frame for $N_{joints}=24$ SMPL joints). This output represents the initial real-time skeletal tracking of the human.

\textbf{Smoothing}
The raw 3D joint coordinates stream is then fed into our Human Pose Sequence Tracking Model (HPSTM), implemented as the \textit{Smoothing} module.
\begin{itemize}
    \item \textbf{Buffering \& Windowing:} For real-time processing, HPSTM maintains an internal sliding window buffer of the $W$ most recent frames (where $W$ is the \textit{window\_size}). As new frames arrive, they are added to the buffer, and the oldest frames are discarded, allowing for continuous processing of a fixed-length sequence.
    \item \textbf{Sequence Processing \& Model Adaptation:} The \textit{Transformer} model within HPSTM processes this entire temporal window. It predicts key \textbf{pose parameters} for each frame, including root joint orientation as a quaternion, local joint rotations for all other joints relative to their parents, and per-segment bone lengths. These parameters define the pose on a human skeletal manifold. A differentiable Forward Kinematics (FK) module reconstructs the refined 3D joint positions from these predicted parameters, ensuring anatomical consistency. The model leverages the temporal context within the window to smooth pose data, filter noise, and handle occlusions. Furthermore, the prediction of Cholesky L factors for covariance matrices associated with each joint's refined 3D position supports a probabilistic approach, representing uncertainty in the pose estimates.
    \item \textbf{Output:} HPSTM outputs a refined sequence of 3D joint coordinates for the entire window (shape: $N_{joints} \times 3 \times W$) and the associated Cholesky L factors for the covariance matrices (shape: $N_{joints} \times 3 \times 3 \times W$).
\end{itemize}

\textbf{Filtering \& Mapping}
The final stage, \textit{skeleton-to-uarm} module, maps the refined human motion to the robotic arm.
\begin{itemize}
    \item \textbf{Input \& Dynamic Loading:} This module receives the entire refined sequence (the $N_{joints} \times 3 \times W$ array of coordinates) from HPSTM's latest processed window. To achieve fluid motion, this module maintains its own internal buffer, which is dynamically updated with these incoming smoothed pose sequences. The mapper consumes frames from this buffer, and then to iterate through pre-smoothed frames and send a continuous stream of commands. This sequence-based handoff is crucial for mitigating choppiness.
    \item \textbf{Processing:} The controller module extracts relevant arm joint data (e.g., right shoulder, elbow, wrist) based on configurations. Then pose vector calculator maps the human arm's relative pose (wrist relative to shoulder) to the uArm's coordinate system, involving dynamic scaling, coordinate transformations, and workspace clipping.
    \item \textbf{uArm Control \& Output:} Calculated target uArm coordinates (X, Y, Z) and wrist angles are sent as non-blocking commands to the uArm Swift Pro via its SDK. These commands populate the uArm's internal buffer, facilitating continuous motion.
\end{itemize}

\section{Results}
\label{sec:results}
The overall data flow is: Video Frame $\rightarrow$ \textit{romp\_tracking} $\rightarrow$ Raw 3D Coords $\rightarrow$ HPSTM (Windowed Refinement) $\rightarrow$ Refined 3D Coords Sequence \& Covariance Factors $\rightarrow$ \textit{skeleton-to-uarm} (Dynamic Buffering \& Mapping) $\rightarrow$ uArm Commands $\rightarrow$ Physical Robot Motion. This integrated pipeline, particularly the role of HPSTM in refining motion data into smoothed sequences with uncertainty quantification, is designed to significantly enhance the smoothness, naturalness, and potential robustness of the mimicry.

We evaluated our proposed Human Pose Sequence Tracking Model (HPSTM) against several baselines on 3D human pose sequences from the real capture pose. The input sequences were corrupted with a combination of Gaussian noise, bone-length jitter, temporal jitter, and outliers to simulate realistic noisy conditions. We report performance using metrics for accuracy, smoothness, and physical plausibility. The primary baselines include the noisy input itself (Noisy Input), a simple Particle Filter (PF Smoothed)\footnote{The PF baseline, employing a constant velocity model, exhibited significant divergence, yielding MPJPE values several orders of magnitude higher than other methods (e.g., $>$100mm), and is thus not considered a competitive baseline for detailed comparison in accuracy.}, and a Savitzky-Golay filter (SavGol Smoothed). We compare two variants of our HPSTM: `Old HPSTM` (predicts pose parameters including bone lengths, without covariance estimation) and `New HPSTM` (additionally predicts per-joint covariance).

\subsection{Quantitative Evaluation}
The quantitative results for a representative test sequence with a mixed noise profile comprising Gaussian displacement ($\sigma=0.03$\,m), filtered temporal jitter (signal $\sigma_t=0.03$\,m, $w_t=7$), bone-length perturbation (relative $\sigma_{bl}=8\%$), and outlier corruption ($p_{out}=0.25\%$, max dev. $s_{out}=0.25$\,m) are presented in Table~\ref{tab:quantitative_results}.

\begin{table}[htbp]
\centering
\caption{Quantitative comparison of pose smoothing methods. Accuracy metrics (MPJPE, PA-MPJPE, RR-MPJPE, BoneMAE) are in mm. Lower is better for all metrics except where noted.}
\label{tab:quantitative_results}
\resizebox{\columnwidth}{!}{%
\begin{tabular}{lccccc}
\toprule
Metric                & Noisy Input & PF Smoothed & SavGol Smoothed & Old HPSTM & New HPSTM \\
\midrule
 MPJPE (mm)            & 54.4312     & 143.8784    & 25.7914         & 33.6487   & 37.5019   \\
 PA-MPJPE (mm)         & 56.1329     & 160.2433    & 26.4773         & 36.6353   & 39.0942   \\
 RR-MPJPE (mm)         & 72.6348     & 178.1533    & 34.4605         & 36.7478   & 40.3403   \\
 MeanAccel             & 0.1259      & 0.0064      & 0.0111          & 0.0009    & 0.0008    \\
 MeanJerk              & 0.2294      & 0.0113      & 0.0185          & 0.0007    & 0.0005    \\
 BoneMAE (mm)          & 49.8573     & 85.4329     & 43.3986         & 35.8387   & 28.6887   \\
 BoneStdDev (mm)       & 43.7854     & 76.9262     & 21.1238         & 2.2541    & 1.6750    \\
 \bottomrule
 \end{tabular}%
 }
 \end{table}

\textbf{Accuracy.}  
On the hardest corruption benchmark Savitzky–Golay (SG) attains the
lowest MPJPE at \textbf{25.79 mm}, whereas HPSTM-Old and HPSTM-New reach
33.65 mm and 37.50 mm, respectively, and the raw noisy input remains at
54.43 mm.  PA-MPJPE and RR-MPJPE show the same ordering, confirming
SG’s superior frame-wise fit under extreme noise.

\textbf{Smoothness.}  
Both HPSTM variants produce markedly steadier motion than SG.  
HPSTM-New reports a MeanAccel of \textbf{$8\times10^{-4}$} and MeanJerk
of \textbf{$5\times10^{-4}$}, HPSTM-Old follows closely
($9\times10^{-4}$, $7\times10^{-4}$), while SG is an order of magnitude
higher (0.0111, 0.0185).  Hence HPSTM delivers low-jerk trajectories
critical for downstream control and animation.

\textbf{Physical plausibility.}  
HPSTM also best preserves bone structure.  
HPSTM-New reduces BoneMAE to \textbf{28.69 mm} and BoneStdDev to
\textbf{1.68 mm}, outperforming SG (43.40 mm, 21.12 mm) and the noisy
input (49.86 mm, 43.79 mm).  HPSTM-Old achieves 35.84 mm / 2.25 mm,
still far tighter than the baselines, indicating that the
forward-kinematics manifold and bone-length head successfully enforce
anatomical consistency.

\subsection{Different Hyper-parameter Compare}
\label{sec:hyper_ablation}
To highlight the effect of curriculum noise and covariance learning, we
report a compact ablation across eight training schedules in
Table~\ref{tab:validation_results}.  The rows are organised so that each
clean-data model (Cfg.~1, 3, 5, 7) is immediately followed by its
noise-augmented counterpart (Cfg.~2, 4, 6, 8); columns span joint
accuracy, temporal smoothness, and bone-length integrity.  This layout
allows the reader to (i) compare the impact of noise injection
horizontally within each pair, and (ii) assess the cost–benefit of
enabling covariance prediction vertically across the ``No Cov’’ and
``+Cov’’ blocks under both evaluation regimes (\emph{No Noise} vs.\
\emph{Complex Noise}).  Key trends distilled from the table are
summarised in \S\ref{sec:discussion}.

\textbf{Accuracy.}  
\emph{Clean evaluation:} the ``Noise + No Cov'' variant (Cfg--2) reaches the lowest MPJPE at \textbf{25.99\,mm}, an 18.6\,\% improvement over the clean-only baseline (Cfg--1, 31.94\,mm).
\emph{Complex Noise evaluation:} the same recipe (Cfg--6) again leads with \textbf{27.34\,mm}, reducing error by 20.6\,\% relative to its clean-only counterpart (Cfg--5, 34.44\,mm).
Adding covariance consistently raises positional error, e.g.\ Cfg--1 $\rightarrow$ Cfg--3 (31.94\,mm $\rightarrow$ \textbf{36.09\,mm}) and Cfg--6 $\rightarrow$ Cfg--8 (27.34\,mm $\rightarrow$ \textbf{33.54\,mm}).

\textbf{Smoothness.}  
All models keep jerk well below $10^{-3}$.  
The best scores arise from the covariance-enabled, noise-trained setting under Complex Noise (Cfg--8):  
MeanAccel = \textbf{$5\times10^{-4}$}, MeanJerk = \textbf{$7\times10^{-4}$}.

\textbf{Physical plausibility.}  
Covariance prediction sharpens bone consistency on clean data: BoneMAE falls from 35.42\,mm (Cfg--1) to \textbf{28.54\,mm} (Cfg--3); BoneStdDev from 2.02\,mm to \textbf{1.47\,mm}.
In noisy training, the bone-length benefit persists (Cfg--6 $\rightarrow$ Cfg--8: BoneStdDev 1.94\,mm $\rightarrow$ \textbf{1.90\,mm}) while MPJPE rises.

\begin{figure}[htbp]
  \centering
  \includegraphics[width=1\linewidth]{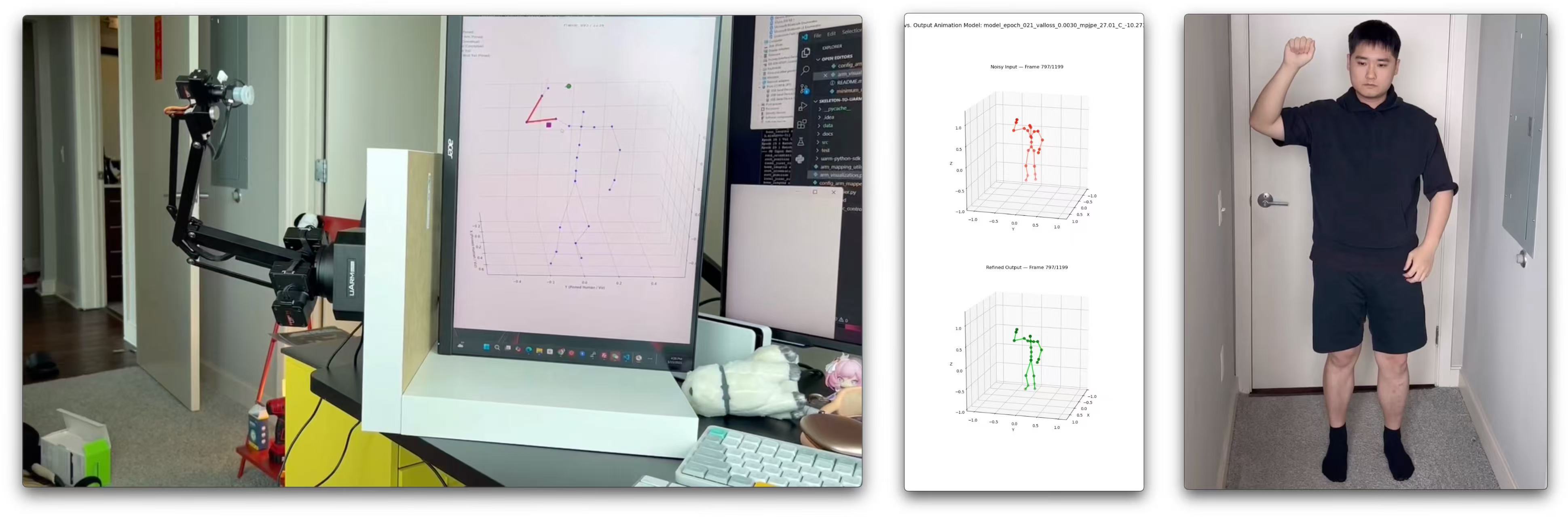}
  \caption{Experiment Screenshot}
  \label{fig:frames}
\end{figure}

\begin{table*}[!t]
  \caption{Validation performance after 30 epochs under different training configurations. 
           ``Noise’’ indicates noisy data used from epoch~11; 
           ``Covariance’’ indicates covariance prediction enabled from epoch~21.}
  \label{tab:validation_results}
  \centering
  \setlength{\tabcolsep}{4pt}
  \scriptsize
  \begin{tabular}{@{}lrrrrrrrr@{}}
\toprule
Experiment Configuration & Noise & MPJPE\,(mm) & PA-MPJPE\,(mm) & RR-MPJPE\,(mm) & MeanAccel & MeanJerk & BoneMAE\,(mm) & BoneStdDev\,(mm)\\
\midrule
1.~Clean Data, No Covariance            & No Noise      & 31.943 & 34.123 & 35.623 & 0.0003 & 0.0002 & 35.420 & 2.017 \\
2.~Noise (ep.~11), No Covariance        & No Noise      & 25.986 & 27.812 & 27.419 & 0.0005 & 0.0003 & 33.302 & 1.896 \\
3.~Clean, +Covariance (ep.~21)          & No Noise      & 36.094 & 37.659 & 39.448 & 0.0004 & 0.0003 & 28.536 & 1.473 \\
4.~Noise (ep.~11), +Covariance (ep.~21) & No Noise      & 32.316 & 35.286 & 32.120 & 0.0004 & 0.0002 & 35.047 & 1.524 \\
\addlinespace
5.~Clean Data, No Covariance            & Complex Noise & 34.435 & 37.001 & 37.399 & 0.0008 & 0.0007 & 36.164 & 2.564 \\
6.~Noise (ep.~11), No Covariance        & Complex Noise & 27.338 & 29.273 & 28.493 & 0.0007 & 0.0007 & 33.459 & 1.936 \\
7.~Clean, +Covariance (ep.~21)          & Complex Noise & 38.065 & 40.493 & 40.961 & 0.0009 & 0.0007 & 28.567 & 1.889 \\
8.~Noise (ep.~11), +Covariance (ep.~21) & Complex Noise & 33.543 & 36.660 & 33.358 & 0.0005 & 0.0007 & 34.908 & 1.895 \\
\bottomrule
\end{tabular}
\end{table*}

\section{Discussion}
\label{sec:discussion}

\textbf{Key trade-off.}  
HPSTM shifts the optimisation target from \emph{minimum distance} to \emph{physically valid smooth motion}.  
Compared with a Savitzky–Golay baseline, it reduces BoneMAE and BoneStdDev by 30–50 \% while preserving sub-millimetre MeanAccel/MeanJerk, at the cost of a modest MPJPE increase.  For tasks where skeletal integrity and temporal stability are critical (e.g.\ robot control), this compromise is acceptable.

\textbf{Synthetic noise acts as a robust regulariser.}  
Adding corruption from epoch 11 (cfg.~2, 4, 6, 8) yields a consistent MPJPE drop of 18–20 \% on clean data and 5–8 \% under the hardest “Complex Noise’’ test, confirming the value of curriculum noise exposure.

\textbf{Covariance prediction improves bone consistency but can hurt position error.}  
With clean training data, enabling the covariance head cuts BoneMAE from 35.4 mm to 28.5 mm (cfg.~1 → 3) yet raises MPJPE to 36.1 mm.  When noise is already present (cfg.~2 → 4, 6 → 8) the bone-length gain diminishes while the MPJPE penalty remains, indicating a competition between uncertainty modelling and point accuracy.

\textbf{Best-performing setting.}  
For overall accuracy under realistic noise, \emph{Noise\,\,+\,No Covariance} (cfg.~6) achieves the lowest MPJPE (27.3 mm) and the smallest relative degradation ( +5.2 \%) from clean to noisy testing.

\textbf{Implication.}  
Practitioners may choose between (i) training with noise only for best joint accuracy, or (ii) adding covariance to secure anatomical plausibility when bone fidelity is paramount.  Future work will explore adaptive loss weighting and uncertainty-aware control that fully exploit HPSTM’s covariance outputs.

A particularly promising avenue for future work lies in advancing the pose-mapping module by leveraging learned approaches such as Action Chunking Transformers (ACT) and action tokens. Moving beyond the current geometric mapping, an ACT-based system could learn a more nuanced and robust translation of smoothed human arm motion from HPSTM to the robotic arm's workspace. By training on paired human-robot motion data, this approach could capture temporal abstractions and better navigate complex kinematic discrepancies, potentially enabling the robot to understand human movement intent and execute more natural, intelligent, and context-aware imitations, especially for higher-DoF robots or more intricate tasks.

\section{Acknowledgment}

The authors would like to express their sincere gratitude
to Professor Pratik Chaudhari for his generous hardware
support, specifically for providing the robotic arm used in
our experiments. This support was invaluable to the successful
completion of this research.

\end{document}